\documentclass[12pt]{article}

\usepackage{amsmath}
\usepackage{amsthm}
\usepackage{times}
\usepackage{graphicx}
\usepackage{color}
\usepackage{multirow}
\usepackage{subfig}

\newtheorem{myDef}{Definition}

\begin{document}
{\LARGE Knowledge Forest: A Novel Model to Organize Knowledge Fragments}

\ \\
{\bf Qinghua Zheng$^{\displaystyle 1,2}$}\\
{qhzheng@mail.xjtu.edu.cn}\\
{\bf Jun Liu$^{\displaystyle 1,2}$}\\
{liukeen@mail.xjtu.edu.cn}\\
{\bf Hongwei Zeng$^{\displaystyle 1,2}$}\\
{zhw1025@stu.xjtu.edu.cn}\\
{\bf Bei Wu$^{\displaystyle 1,2}$}\\
{wubei784872134@stu.xjtu.edu.cn}\\
{\bf Zhaotong Guo$^{\displaystyle 1,2}$}\\
{zhaotongguo@stu.xjtu.edu.cn}\\
{\bf \large Bifan Wei$^{\displaystyle 2,3}$}\\
{weibifan@xjtu.edu.cn}\\
{$^{\displaystyle 1}$School of Electronic and Information Engineering, Xi'an Jiaotong University, Xi'an, Shaanxi, 710049, China}\\
{$^{\displaystyle 2}$National Engineering Lab for Big Data Analytics, Xi'an Jiaotong University, Xi'an, Shaanxi, 710049, China}\\
{$^{\displaystyle 3}$School of Continuing Education, Xi'an Jiaotong University, Xi'an, Shaanxi, 710049, China}\\

\begin{center} {\bf Abstract} \end{center}
With the rapid growth of knowledge, it shows a steady trend of knowledge fragmentization. Knowledge fragmentization manifests as that the knowledge related to a specific topic in a course is scattered in isolated and autonomous knowledge sources~\cite{liu2015faceted}. We term the knowledge of a facet in a specific topic as a knowledge fragment. 
For example, \textit{``A push operation adds an item to the top-most location on the stack.''} is a knowledge fragment about facet \textit{operation} of topic \textit{Stack}. 
The problem of knowledge fragmentization brings two challenges:
First, knowledge is scattered in various knowledge sources, which exerts users' considerable efforts to search for the knowledge of their interested topics, thereby leading to information overload \cite{huang2013webevis}. 
Second, learning dependencies which refer to the precedence relationships between topics in the learning process are concealed by the isolation and autonomy of knowledge sources, thus causing learning disorientation \cite{bahaha2018smart}.
However, three mainstream knowledge organization models \cite{hodge2000systems,freebase,probase,KnowledgeVault}, including term list, categorization, and relation list, which organize knowledge fragments without facet hyponymy and ignore learning dependencies between topics, can hardly be applied to address these two challenges.

To solve the knowledge fragmentization problem, we propose a novel knowledge organization model, knowledge forest, which consists of facet trees and learning dependencies. Facet trees can organize knowledge fragments with facet hyponymy to alleviate information overload. Learning dependencies can organize disordered topics to cope with learning disorientation. 
The knowledge forest use Resource Description Framework (RDF) for knowledge representation and storage. 
Compared with RDF, knowledge forest organizes knowledge fragments in a way that is more consistent with human cognition and learning.
Furthermore, we propose an effective construction method of knowledge forest. 
The construction process of knowledge forest contains facet tree construction, learning dependency extraction, and knowledge fragment assembly. 

\begin{figure*}
	\centering
	\subfloat[\label{fig.1.a}]{\includegraphics[width=0.30\linewidth]{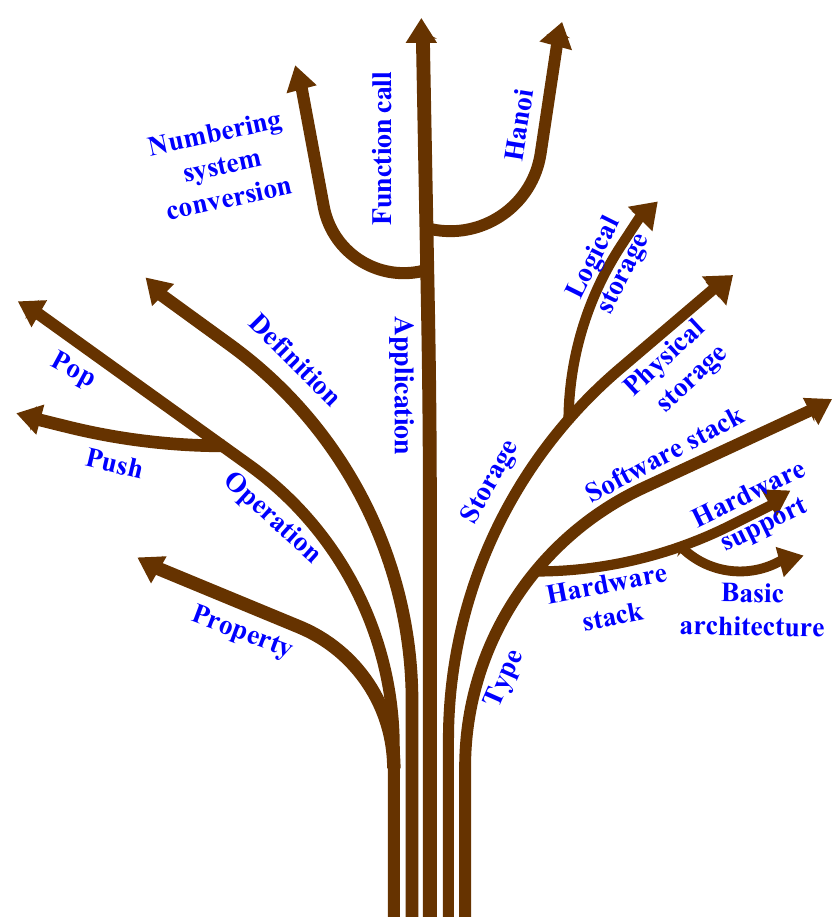}\centering}
	\hspace{90pt}
	\subfloat[\label{fig.1.b}]{\includegraphics[width=0.40\linewidth]{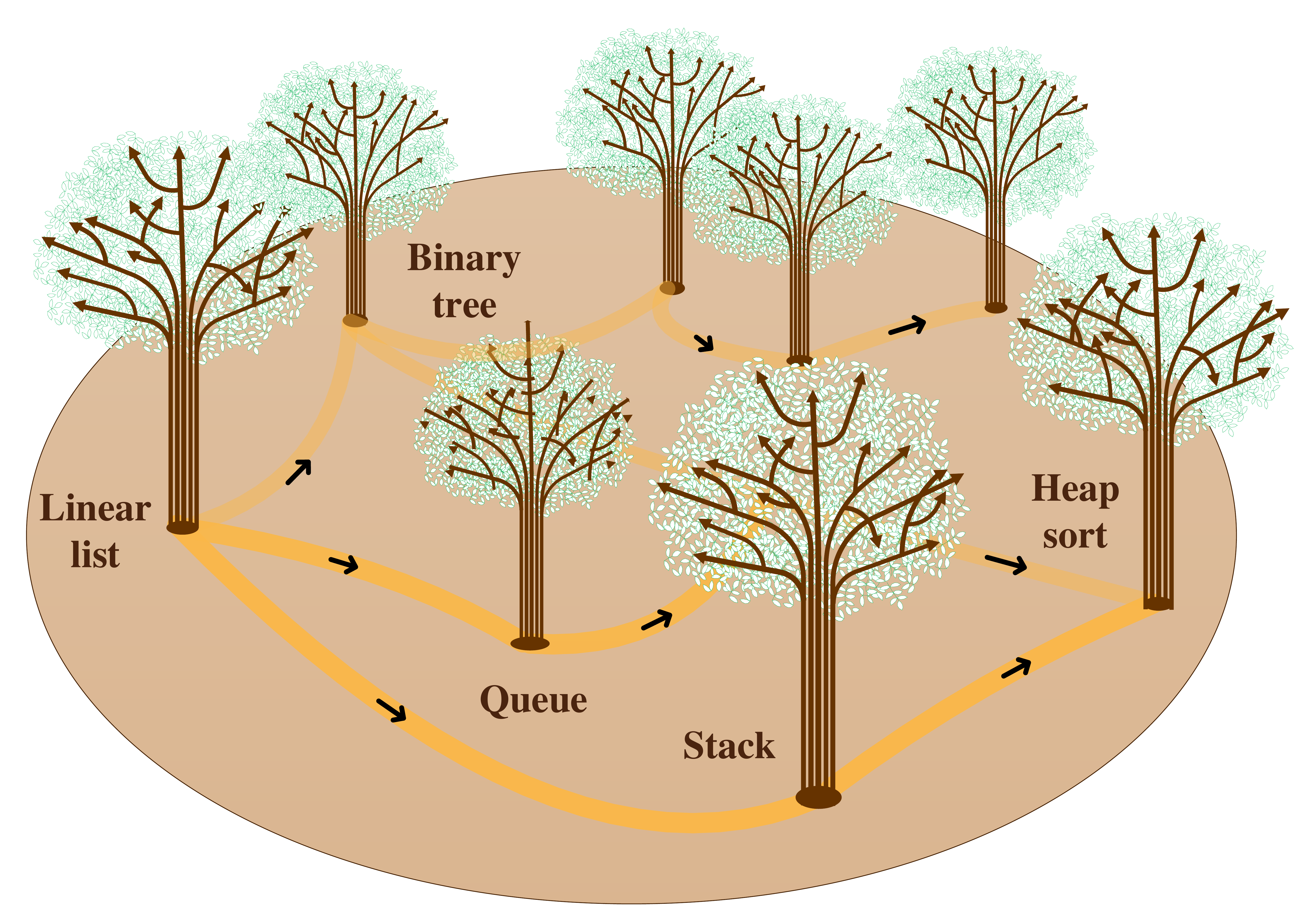}\centering}
	\caption{Visualization of the knowledge organization model. (a) Facet tree of topic \textit{Stack} and (b) the partial view of the knowledge forest of course \textit{Data Structure}.} 
	\label{fig:EcUND} 
\end{figure*}

\section{Definitions and Notations}
The related definitions and formalized notations of knowledge forest are introduced as follows.

\begin{myDef}[Facet Tree]
	A \textbf{Facet Tree} is a set of facets with facet hyponymy. 
	Suppose $ T = \{t_1, ..., t_n\}$ is the topic set of a course, 
	the facet tree of topic $t_i \in T$ can be expressed as a tuple $ FT_{i} = (F_{i}, RF_{i}) $. 
	$ F_{i} $ refers to the facet set corresponding to $ t_i $. 
	$ RF_{i} \subseteq ( \{t_i\} \cup F_{i} ) \times F_{i} $ represents topic-facet and facet-facet relationships. 
	For example, Figure \ref{fig.1.a} shows the facet tree of topic \textit{Stack}, 
	$F_{\textit{Stack}} = \{ \textit{storage}, \textit{operation}, \textit{pop}, \textit{push}, \dots \}$ and $RF_{\textit{Stack}} = \{(\textit{Stack}, \textit{operation}), (\textit{operation}, \textit{pop}), \dots \}$.
\end{myDef}

\begin{myDef}[Materialized Facet Tree]
	A \textbf{Materialized Facet Tree} is a facet tree which is assembled with knowledge fragments. 
	The materialized facet tree of topic  $ t_i \in T $ can be expressed as a triple  $ MFT_{i} = (FT_{i}, K_{i}, FK_{i}) $. 
	$ K_{i} $ refers to the set of knowledge fragments corresponding to topic ${ t_i }$. 
	$ FK_{i} \subseteq F_{i} \times K_{i} $ is the mapping relationship set between facet set $ F_{i} $ and knowledge fragment set $ K_{i} $. 
	Knowledge fragment $k \in K_{i}$ will be assembled to facet tree $ FT_{i} $ according to mapping relationship $ FK_{i} $. 
	For example, knowledge fragment \textit{``A push operation adds an item to the top-most location on the stack.''} will be assembled to facet \textit{operation} of topic \textit{Stack}. 
\end{myDef}

\begin{myDef}[Knowledge Forest]
	A \textbf{Knowledge Forest} is the combination of materialized facet trees of topics and learning dependencies between topics. 
	The knowledge forest can be expressed as a tuple $ KF = (MFT, LD)$. 
	$ MFT = \{MFT_{i} \  | \  t_i \in T \} $ refers to the set of materialized facet trees corresponding to all topics in $T$. 
	$ LD \subseteq T \times T $ represents learning dependencies between topics in $T$. 
	Figure \ref{fig.1.b} is a partial view of knowledge forest of course \textit{Data Structure}. 
	The relationship with an arrow represents the learning dependency. For example, the learning dependency from topic \textit{Linear list} to \textit{Stack} indicates that we should learn \textit{Linear list} first. 
\end{myDef}

\section{Knowledge Forest Construction}
The construction process of knowledge forest, which can be seen as the organization process of knowledge fragments, contains three steps as follows.  

\textbf{Step 1.} We propose a facet propagation algorithm to construct facet trees. The method aims to give a facet set $F_i$ for each topic $t_i \in T$. Intuitively, parent-child topic pair and brother topic pair both have similar facet sets. The parent-child topic pair includes two topics with hypernymy relationship. The brother topic pair includes two topics whose hypernym topic is the same one. 
For each topic, we parse and preprocess the \textit{Contents} section of the corresponding Wikipedia webpage\footnote{\url{https://en.wikipedia.org/}} to obtain the initial facet set. 
Then, we use the facet propagation algorithm to complete the facet set of each topic on the basis of the initial facet set. 
During each facet propagation, the probability of facet $f \in F_i$ will be updated by the facet set similarity between both parent-child topic pairs and brother topic pairs. 
Until the algorithm converges, the probability of $f \in F_i$ bigger than 0.5 indicates that topic $t_i$ includes facet $f$, and vice versa.

\textbf{Step 2.} We utilize our early work \cite{liu2011mining} to extract learning dependencies among topics. 
This method proposes two useful hypotheses, the distribution asymmetry of core terms and the locality of learning dependencies, which are essential for building the classification model to identify learning dependencies. 

\textbf{Step 3.} We propose a mapping method based on Convolutional Neural Network (CNN) to assemble knowledge fragments to corresponding facet tree \cite{Wu2018Facet}. 
This method aims to give one or more facet labels $F'_i\subseteq F_i$ for each $k \in K_i$, which consists of three steps. 
(i) We employ word embeddings to represent the words of knowledge fragments. 
Then, we use three convolution layers and three pool layers to represent each knowledge fragment as three matrices indicating the phrases information, 
corresponding to \textit{unigram}, \textit{bigram} and \textit{trigram}, respectively. 
(ii) 
To reduce facet heterogeneity, 
we propose a text matching strategy to establish the relationship between each knowledge fragment and a Facet Label Text (FaLT).
First, we introduce FaLTs from Wikipedia webpages. 
For example, the FaLT corresponding to facet \textit{definition} of topic \textit{Stack} is ``\textit{In computer science, a stack is an abstract data type that serves as a collection of elements, with two principal operations}\footnote{https://en.wikipedia.org/wiki/Stack (abstract\_data\_type)}.''
Then, each FaLT is represented as three matrices by the knowledge fragment representation method mentioned above. 
Finally, three-dimensional similarity matrices are generated by cosine similarity measures between a knowledge fragment and a FaLT.
(iii) We utilize the three-dimensional similarity matrices as the input of a three-channel CNN
as multiple binary classifications for facet label assignment.

\section{Datasets and Basic Statistic}
We recruit ten participants who major in computer science with enough knowledge to annotate the knowledge forest. They independently annotate three courses, including \textit{Data Structure}, \textit{Data Mining}, and \textit{Computer Network}. 
Course \textit{Data Structure} contains 193 topics, 35,076 knowledge fragments, and 247 learning dependencies. 
Course \textit{Data Mining} contains 93 topics, 12,723 knowledge fragments, and 128 learning dependencies. 
Course \textit{Computer Network} contain 84 topics, 13,081 knowledge fragments, and 113 learning dependencies.

\section{Experiments}
To validate the effectiveness of automatic construction method of knowledge forest, we conduct experiments on the three courses. 
The $ nDCG $ score of facet tree construction can achieve more than 82\%, and the $Macro\_F$ value of knowledge fragment assembly method can reach more than 83\% on all three courses.
The results indicate that our method implement a good generalization capability and can effectively organize knowledge fragments in different courses. 

To evaluate the effectiveness of knowledge forest to alleviate information overload and learning disorientation, we conduct learning performance test. 
We recruit sixty participants for the three courses, and each course has twenty participants, ten of which in the control group and the other ten in the experimental group. 
We develop a prototype knowledge forest system named Yotta.
The baseline is the control group which do not use Yotta to learn corresponding courses. 
The comparison metric is the \textit{Mean} and \textit{Standard deviation} of participant' scores in pre-test and post-test.
The Student's t-test is used for statistical analysis which can be summarized as follows. 
Firstly, the scores of pre-test, which are concerned with the participants' prior knowledge, have no significant differences between the control group and the experimental group in the three courses $(p>0.05)$. 
Secondly, participants' scores in post-test have significant improvements over pre-test both in the control group and the experimental group $(p<0.05)$. 
Thirdly, the gain scores of the experimental group are much higher than those in the control group, 
which indicates that the participants in the experimental group can achieve significantly better learning performance than those in the control group $(p<0.05)$. 
Thereby, we can conclude that the knowledge forest is useful to alleviate the participants' information overload and learning disorientation.

\section{Conclusion}
In this letter, we propose a novel knowledge organization model, knowledge forest, which consists of facet trees and learning dependencies. 
We propose an automatic construction method of knowledge forest.
The results of extensive experiments conducted on three courses show that knowledge forest can effectively organize knowledge fragments and alleviate information overload and learning disorientation.

\section*{Acknowledgment}
	This work was supported by National Key Research and Development Program of China (2018YFB1004500), National Natural Science Foundation of China (61532015, 61532004, 61672419, 61672418), Innovative Research Group of the National Natural Science Foundation of China (61721002), Innovation Research Team of Ministry of Education (IRT\_17R86), Project of China Knowledge Centre for Engineering Science and Technology.

\end{document}